\documentclass[conference]{IEEEtran}
\IEEEoverridecommandlockouts
\IEEEpubid{\makebox[\columnwidth]{978-1-5386-5541-2/18/\$31.00~\copyright2020 IEEE \hfill} \hspace{\columnsep}\makebox[\columnwidth]{ }}
\usepackage{cite}
\usepackage{amsmath,amssymb,amsfonts}
\usepackage{algorithmic}
\usepackage{graphicx}
\usepackage{textcomp}
\usepackage{xcolor}
\def\BibTeX{{\rm B\kern-.05em{\sc i\kern-.025em b}\kern-.08em
    T\kern-.1667em\lower.7ex\hbox{E}\kern-.125emX}}
\begin{document}

\title{Automated machine vision enabled detection of movement disorders from hand drawn spirals}

\author{\IEEEauthorblockN{Nabeel Seedat}
\IEEEauthorblockA{
\textit{University of the Witwatersrand  \& Shutterstock}\\
South Africa \& USA \\
seedatnabeel@gmail.com}
\and
\IEEEauthorblockN{Vered Aharonson}
\IEEEauthorblockA{\textit{Electrical and Information Engineering} \\
\textit{University of the Witwatersrand}\\
 South Africa \\
vered.aharonson@wits.ac.za}
\and
\IEEEauthorblockN{Ilana Schlesinger
}
\IEEEauthorblockA{\textit{Department of Neurology} \\
\textit{Rambam Health Care Campus}\\
Israel\\
i\_schles@rambam.health.gov.il}

}

\maketitle

\begin{abstract}
A widely used test for the diagnosis of Parkinson’s disease (PD) and Essential tremor (ET) is hand-drawn shapes, where the analysis is observationally performed by the examining neurologist. This method is subjective and is prone to bias amongst different physicians. Due to the similarities in the symptoms of the two diseases, they are  often misdiagnosed. Studies which attempt to automate the process typically use digitized input, where the tablet or specialized equipment are not affordable in many clinical settings. This study uses a dataset of scanned pen and paper drawings and a convolutional neural network  (CNN) to perform classification between PD, ET and control subjects. The discrimination accuracy of PD from controls was 98.2\%. The discrimination accuracy of PD from ET and from controls was 92\%. An ablation study was conducted and indicated that correct hyper-parameter optimization can increases the accuracy up to 4.33\%. Finally, the study indicates the viability of using a CNN-enabled machine vision system to provide robust and accurate detection of movement disorders from hand drawn spirals.
\end{abstract}

\begin{IEEEkeywords}
convolutional neural network, deep learning, Essential Tremor, hyper-parameter optimization, movement disorder, Parkinson's disease
\end{IEEEkeywords}

\section{Introduction}
The usage of machine learning for automated disease diagnosis in an efficient and accurate manner has the potential to reduce labor and cost in healthcare, and improve patient care. Deep learning has been widely studied for automated and quantitative disease assessment particularly for medical imaging, where convolutional neural networks (CNNs) were employed \cite{dl1,dl2}. Other medical applications are typically less studied, mainly since datasets in these other contexts are not as readily available as imaging datasets.

An area in medicine where this problem is prevalent is the quantitative assessment of motor disorders such as Parkinson’s disease (PD) and Essential Tremor (ET). Both these conditions are debilitating and have increasing prevalence \cite{ref1}. Both PD and ET typically manifest in hand tremors which severely impacts the patients' quality of life. Distinguishing between the different types of tremors (PD vs ET) is critical for correct treatment and long term management of the disease \cite{ref2}. The Static Spiral Test (SST) is a widely used test in tremor diagnosis \cite{ref3}. 

This simple and short test requires the patient to retrace Archimedean spirals on paper using a pen. An expert neurologist performs an observational assessment of the subject as they carry out the task, as well as a visual analysis of the drawn spiral. Healthcare professionals and patients could greatly benefit from an automation of the SST since observation-based data is qualitative and subjective, and hence are prone to bias and inaccuracy, which may lead to incorrect treatment. To the best of our knowledge there is no known human baseline for PD, ET and Control discrimination. Additionally, an automated analysis of the SST would allow patients in clinics without expert neurologists to be triaged.

Attempts to automate the SST typically use digitized tablets or instrumented pens embedded with sensors, such as accelerometers and/or pressure sensors, that capture information as the patient draws the spiral \cite{pd1,pd2,pd3}. Although these methods have a potential to capture the relevant motor information from the SST and can easily record quantitative data for analysis, practical clinical considerations often limit their usage. In particular the cost of additional and expensive high-end hardware is often a limitation.

Additionally, most analyses with these devices focus on a binary classification between PD vs controls or ET vs Controls, while movement disorder clinics require a discrimination between PD, ET and controls, since early stage symptoms of PD and ET are mild and may resemble control subjects \cite{ref4}.

This paper therefore makes the following contributions to address these aforementioned limitations:
\begin{itemize}
  \item Discrimination between PD, ET and Controls using an end-to-end deep learning solution.
\item Automated analysis of the SST based on conventional pen and paper tests.
\item An ablation study to highlight the performance improvement which can be obtained through correct hyper-parameter optimization of deep neural networks.

\end{itemize}

\begin{figure*}[t]
    \centering
    \includegraphics[width=0.75\linewidth,trim={1cm 10cm 4.5cm 4cm},clip]{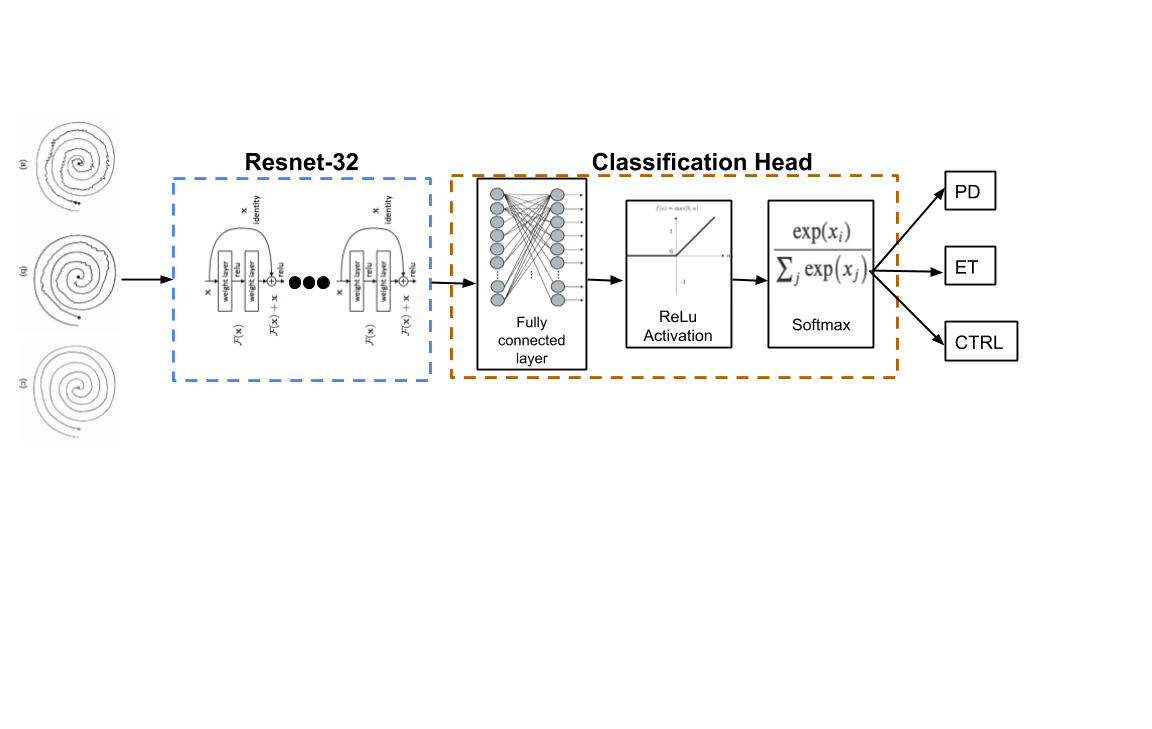}
    \caption{Our proposed end-to-end deep-learning system for the discrimination of Parkinson’s Disease (PD), Essential Tremor (ET) and control patients}
    \label{fig:pipe}
\end{figure*}

\section{Method}
In this work we present a deep-learning based solution applied to the discrimination of Parkinson’s Disease (PD) patients from controls, and PD patients from Essential Tremor (ET) patients and from controls. We propose an end-to-end system illustrated in Figure \ref{fig:pipe}, making use of a convolutional neural network (CNN) to analyze images of the hand drawn SST, hence not requiring any manual feature engineering.

\subsection{Dataset}
\begin{figure*}[htbp]
    \centerline{\includegraphics[width=0.8\textwidth]{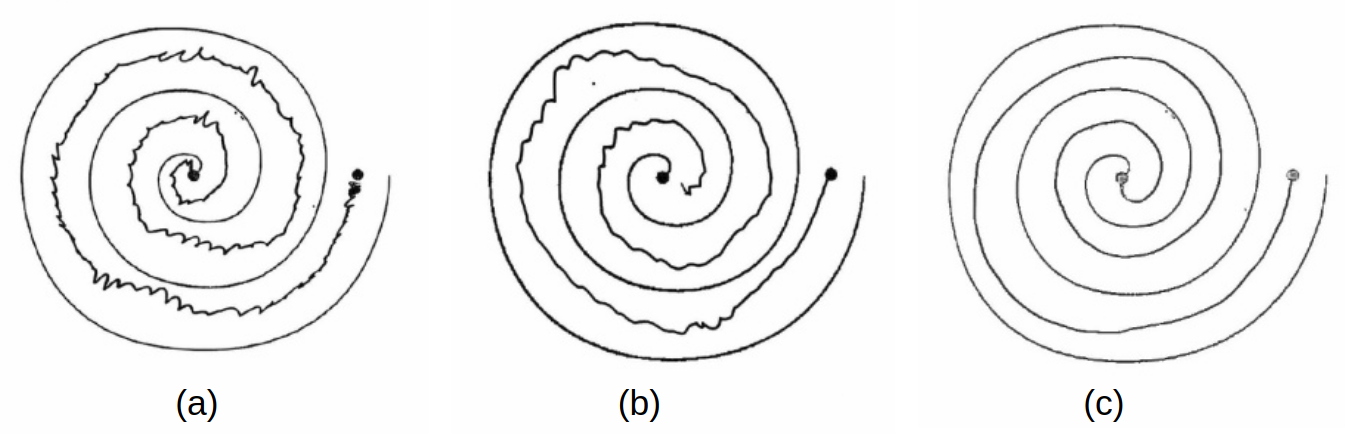}}
    \caption{Typical examples of Static Spiral Test (SST) Archimedean spirals drawn with paper and pen. The three figures represent the three groups involved in this study: (a) Parkinson's disease (PD), (b) Essential Tremor (ET) and (c) Control subject }
    \label{fig:spirals}
\end{figure*}

The dataset consists of camera captured images of hand-drawn Archimedean spirals which were acquired from subjects performing the Static Spiral Test (SST) using a pen and paper. The image dimensions are 300 x 300 x 3.  Examples of the images of the spirals in the three groups studied, Parkinson's disease (PD), Essential Tremor (ET) and controls are illustrated in Figure \ref{fig:spirals}.

The data was acquired during routine neurological assessments in a tertiary hospital and labeled by the examining neurologists. The re-use of the dataset was approved by the hospital as well as the university ethics committee.\\

 The dataset consists of spirals of the following categories:
\begin{itemize}
  \item 370 Parkinson’s disease subjects.
\item  669 Essential Tremor subjects.
\item 357 control subjects.

\end{itemize}

\subsection{End-to-End Deep Learning System}
The proposed end-to-end deep learning system precludes the need for manual feature engineering, with the network learning the underlying feature representations needed for discrimination. Three important sub-modules are important to the overall system and are described below. \\

\subsubsection{Data Augmentation}
Deep neural networks typically require large quantities of data for training. Even in cases where transfer learning is applied, an increased number of samples is beneficial to training. Since changes in orientation and contrast do not directly affect image class as well as the overall perceptive task associated with SST, we apply the following data augmentations to the dataset.\\

\begin{itemize}
\item Random application of a horizontal flip to the image with a probability of 0.5.
\item Random change of image contrast with a probability of 0.1.
\item Random zoom and crop on the image with a probability 0.75.\\

\end{itemize}

\subsubsection{Convolutional Neural Network (CNN) Approach}
We aim to demonstrate the value of a transfer learning approach applied in a biomedical computer vision application, where a pre-trained CNN is used as a base network and is fine-tuned for the specific task. The hypothesis is that the generic features (i.e. edges, gradients) in the earlier CNN layers are still useful representations for the specific SST task, whilst the later layers of the network which learn specific feature representations are re-trained to be representative of the SST classification task at hand. Furthermore, we aim to demonstrate the value of tranfer learning on small datasets (such as the SST dataset) and that it reduces the time to train the CNN, reduces compute resources required compared to training the network from scratch and is still a viable method to attain strong discriminative performance.

Consequently, we make use of a ResNet-32 CNN architecture \cite{resnet} with pre-trained ImageNet weights, which is one of the state of the art networks trained on Imagenet \cite{resnet}. Whilst, other base architectures could be utilized the ResNet (Deep Residual Network) architecture was utilized as the deeper and thinner representation has been shown to provide better generalization which should allow for better translation to the specific task \cite{resnet}. Moreover, the smoother loss surfaces in ResNets allow for easier forward and backward propagation leading to easier optimization when fine-tuning on the task \cite{resnet}.

Finally, the fine-tuned architecture built on top of the pre-trained ResNet, consists of a fully-connected dense layer where the number of outputs for the softmax equals the number of classes classified. Specifics of training the network is discussed under the experimental analysis.\\

\subsubsection{CNN Hyper-parameter optimization}
Neural networks are also highly sensitive to the specific hyper-parameters on which the network is trained. Therefore, we aim to demonstrate the value of hyper-parameter optimization in ensuring optimal classification performance.

In particular, we optimize the learning rate as it directly impacts the magnitude of parameter updates in the network. We apply two techniques that demonstrate the importance of this optimization:\\

     \textit{Cyclical learning rate:} \\As per the work of \cite{clr}, the aim of a cyclical learning rate policy is to allow the traversal of saddle points and local minima in the loss landscape. This is done by varying the learning rate over an epoch between a lower and upper threshold, where the periodic higher learning rate assists in the traversal of the saddles and local minima points. Furthermore, this allows not only for fewer experiments (and by virtue computations) to find optimal learning rates, but also this policy results in superior accuracy when compared to a singular learning rate with decay.\\
    
    \textit{Discriminative learning rate:} \\As per the work of \cite{howard,layer}, the network is divided into groups of layers from earlier to later layers. Earlier groups of layers are trained with a lower learning rate as the weights represent generic features that do not need to be adapted to the task, whilst later layers are trained at a higher learning rate as the weights need to adapted specifically to the classification task. Hence, we divide our network into three weight groups- early, middle and late. We apply an adaptive policy of learning rates to these three weight groups as follows: Early: $10^{-6}$, Middle: $10^{-4}$, Late: $10^{-2}$.

\subsection{Technical pipeline}
Our technical evaluation pipeline then follows the following protocol:
\begin{enumerate}
   \item We perform 5-fold cross-validation, such that during each fold: 80\% of the data is used as training data, whilst 20\% of the data is held-out as an unseen test dataset.
    \item Apply data augmentation as described in Section 2.2 to the training dataset.
    \item Use a Resnet32 CNN pre-trained on Imagenet as the base network. Since the pre-trained Resnet32 has an input image size requirement of 224x224x3, all images are resized using nearest neighbor interpolation.
    \item Remove the final fully-connected layer from the pre-trained network.
    \item Add a dense fully connected layer (Multi Layer Perceptron) with the correct number of outputs in the softmax based on the number of classification classes (two for experiment 1 and three for experiment 2). 
    \item Freeze the pre-trained CNN networks weights and train the dense layers with a high learning rate ($10^{-2}$) for 5 epochs.
    \item Unfreeze the entire networks weights and fine-tune the networks for 3 epochs at a low learning rate using the the three weight groups - Early: $10^{-6}$, Middle: $10^{-4}$, Late: $10^{-2}$.
    \item Repeat the cross-validation process (steps 2-7) to ensure numerical stability/robustness and compute the mean and standard deviation of the 5-fold cross-validation.
\end{enumerate}

\section{Experimental Analysis}
Two sets of experiments are carried out on the hand-drawn SST data. The  first experiment carries out discrimination between PD and control subjects and the second experiment carries out discrimination between PD, ET and control subjects. \\

Both experiments make use of the experimental protocol outlined in Section II (c). An ablation study is carried out for steps 6-7, where steps 6 and 7 are carried out with and without the learning rate hyper-parameter optimization techniques described in Section 2.2. The aim is to demonstrate the performance benefit obtained by making using of these optimizations.

\subsection{Discrimination between Parkinson's disease and control subjects}
This experiment aims to classify between PD subjects and controls based on the SST test images. The results shown in Table 1 are obtained from the 5-fold cross-validation.

\begin{table}[htbp]
\centering
\caption{5-fold cross-validation accuracy of the discrimination between PD and Control subjects with and without learning rate hyper-parameter optimization }
\label{sample-table}

\begin{tabular}{c|c}
\textit{Configuration} & \textit{Accuracy  (\%)} \\ \hline
\textbf{With Hyper-parameter optimization}  & \textbf{98.2 $\pm$ 1.35}      \\ \hline
Without Hyper-parameter optimization   & 95.3 $\pm$ 1.66
\end{tabular}
\end{table}

\subsection{Discrimination between Parkinson's disease, Essential Tremor and controls}
This experiment aims to classify between PD subjects, ET subjects and controls based on the SST test images. The results shown in Table 2 are obtained from the 5-fold cross-validation.

\begin{table}[htbp]
\centering
\caption{5-fold cross-validation accuracy of the discrimination between PD, ET and Control subjects with and without learning rate hyper-parameter optimization over 5 repeated experiments }
\label{sample-table}
\begin{tabular}{c|c}
\textit{Configuration} & \textit{Accuracy  (\%)} \\  \hline
\textbf{With Hyper-parameter optimization}   & \textbf{92 $\pm$ 0.614 }    \\  \hline
Without Hyper-parameter optimization   & 87.67 $\pm$  1.02 
\end{tabular}
\end{table}

\begin{figure}[htbp]
    \centerline{\includegraphics[width=0.485\textwidth]{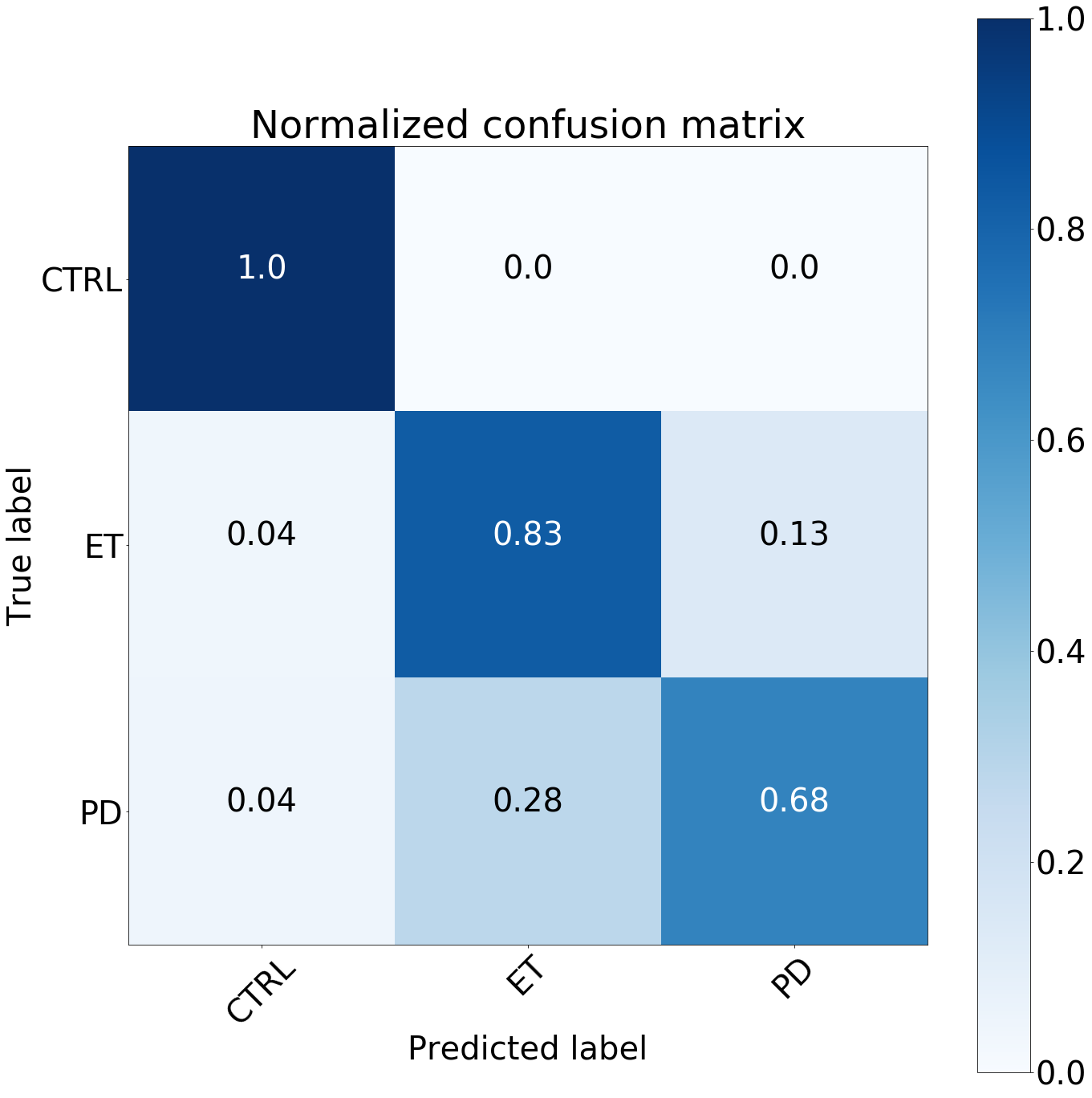}}
    \caption{Normalized confusion matrix of the CNN used to classify between PD, ET and Controls averaged over the 5-fold cross-validation. The x-axis represents the predicted label and the y-axis the true label.}
    \label{fig:res}
\end{figure}

The confusion matrix shown in Figure \ref{fig:res} represents the averaged and normalized 5-fold cross-validation results obtained for the three-class discrimination (PD vs ET vs Control), making use of the hyper-parameter optimizations.

\section{Discussion}
This study presents an automated machine learning discrimination of PD, ET and Controls using images of the hand written SST. This is a preliminary work, which aimed to validate the techniques within this application domain. 

The results convey that an end-to-end deep learning solution using a CNN is both robust and accurate at detecting and discriminating between these three classes in a reliable and autonomous manner, whilst still fitting in with current clinical practices.

The proposed solution where the SST test is conducted using off-the-shelf writing equipment and paper eliminates the need for additional and expensive hardware such as digitized tablets or specialized instrumented writing instruments. This would allow the test to be performed easily in a clinical setting and thus, making it easier to carry out the test in busy clinical settings.

The results imply that the optimal configuration for discrimination both of PD vs Controls and between PD, ET and controls is a ResNet-32 CNN, with the pre-trained ImageNet weights being fine-tuned for the task. The mean  5-fold cross-validation accuracy for the PD vs Control discrimination was 98.2\%, whilst the mean accuracy  for PD, ET and Control discrimination was 92\%. 

In particular, the confusion matrix for the PD, ET and Control discrimination shows that that mis-classification is typically between ET and PD. This is understandable as both PD and ET are movement disorders that result in tremor, which would manifest in the spirals of the subjects. The strength of this result is that even when PD or ET is mis-classified it is still mis-classified as a movement disorder rather than as a control, which is beneficial in a triaging and referral scenario.

The result of the ablation study demonstrates the value of learning rate hyper-parameter optimization techniques. The cyclical learning rate and discriminative learning rate policies combined to increase discrimination accuracy by 4.33\% with no change to the network architecture, thereby highlighting the value of correct hyper-parameter tuning in order to maximize performance of neural networks.

Finally, the value of transfer learning was demonstrated by the overall high discriminative accuracy of the networks. This highlights the value of using pre- trained networks even in biomedical applications and that the high level feature representations from pre-trained networks are useful even on significantly different tasks than the original ImageNet task. Moreover, the benefit is that fewer epochs are required to train the networks which reduces the computational requirements, as well as, training time.

The combined ease of use and machine learning discrimination proposed in this study has the potential to assist healthcare professionals in motor disorder evaluation using the SST, while easily fitting into the clinical environment by making use of the current pen and paper SST. The method could allow healthcare practitioners to easily and quantitatively diagnose motor disorder subjects using standard tools. 

Future work could expand this study and aim at finer-grained classification, seeking to classify the severity, or stage of the motor diseases.

\end{document}